\documentclass[10pt,twocolumn,letterpaper]{article}

\usepackage{cvpr}              

\usepackage{graphicx}
\usepackage{amsmath}
\usepackage{amssymb}
\usepackage{booktabs}

\usepackage{multirow}
\usepackage{enumitem}

\usepackage[accsupp]{axessibility}

\usepackage[pagebackref,breaklinks,colorlinks]{hyperref}

\usepackage[capitalize]{cleveref}
\crefname{section}{Sec.}{Secs.}
\Crefname{section}{Section}{Sections}
\Crefname{table}{Table}{Tables}
\crefname{table}{Tab.}{Tabs.}

\def\cvprPaperID{8237} 
\def\confName{CVPR}
\def\confYear{2023}

\begin{document}

\title{The Resource Problem of Using Linear Layer Leakage Attack in Federated Learning}

\author{Joshua C. Zhao$^1$, Ahmed Roushdy Elkordy$^2$, Atul Sharma$^1$, Yahya H. Ezzeldin$^2$ \\Salman Avestimehr$^2$, Saurabh Bagchi$^1$\\
$^1$Purdue University \\$^2$University of Southern California\\
{\tt\small \{zhao1207,sharm438,sbagchi\}@purdue.edu}, 
{\tt\small \{aelkordy,yessa,avestime\}@usc.edu}
}
\maketitle
\newcommand{\name}{\textsc{Mandrake}\xspace}
\newcommand{\namemeaning}{A magical and sentient plant in the Harry Potter universe that could restore creatures to their original form.}
\newcommand{\atul}[1]{\ifdraft{\textcolor{purple}{Atul: #1}}\fi}
\newcommand{\saurabh}[1]{\ifdraft{\textcolor{red}{Saurabh: #1}}\fi}
\newcommand{\salman}[1]{\ifdraft{\textcolor{blue}{Salman: #1}}\fi}

\begin{abstract}

Secure aggregation promises a heightened level of privacy in federated learning, maintaining that a server only has access to a decrypted {\em aggregate} update. Within this setting, linear layer leakage methods are the only data reconstruction attacks able to scale and achieve a high leakage rate regardless of the number of clients or batch size. This is done through increasing the size of an injected fully-connected (FC) layer. However, this results in a resource overhead which grows larger with an increasing number of clients. We show that this resource overhead is caused by an incorrect perspective in all prior work that treats an attack on an aggregate update in the same way as an individual update with a larger batch size. Instead, by attacking the update from the perspective that aggregation is combining multiple individual updates, this allows the application of sparsity to alleviate resource overhead. We show that the use of sparsity can decrease the model size overhead by over 327$\times$ and the computation time by 3.34$\times$ compared to SOTA while maintaining equivalent total leakage rate, 77\% even with $1000$ clients in aggregation.
\end{abstract}

\vspace{-18 pt}
\section{Introduction}
\label{sec:introduction}
\vspace{-1mm}
Federated learning (FL)~\cite{mcmahan2017communication} has been hailed as a privacy-preserving method of training. 
FL involves multiple clients which train their model on their private data before sending the update back to a server. The promise is that FL will keep the client data private from all (server as well as other clients) as the update cannot be used to infer information about client training data.

However, many recent works have shown that client gradients are not truly privacy preserving. Specifically, data reconstruction attacks~\cite{fowl2022robbing,boenisch2021curious,yin2021see,geiping2020inverting,kariyappa2022cocktail,zhu2019deep,pasquini2021eluding,wen2022fishing} use a model update to directly recover the private training data. These methods typically consist of gradient inversion~\cite{yin2021see,geiping2020inverting,zhu2019deep} and analytic attacks~\cite{boenisch2021curious,fowl2022robbing,kariyappa2022cocktail,lam2021gradient,pasquini2021eluding,wen2022fishing}. Gradient inversion attacks observe an honest client gradient and iteratively optimizes randomly initialized dummy data such that the resulting gradient becomes closer to the honest gradient. The goal is that dummy data that creates a similar gradient will be close to the ground truth data. These methods have shown success on smaller batch sizes, but fail when batch sizes become too large. Prior work has shown that reconstruction on ImageNet is possible up to a batch size of 48, although the reconstruction quality is low~\cite{yin2021see}. Analytic attacks cover a wide range of methods. Primarily, they use a malicious modification of model architecture and parameters~\cite{wen2022fishing, pasquini2021eluding}, linear layer leakage methods~\cite{fowl2022robbing,boenisch2021curious}, observe updates over multiple training rounds~\cite{lam2021gradient}, or treat images as a blind-source separation problem~\cite{kariyappa2022cocktail}. However, most of these approaches fail when secure aggregation is applied~\cite{bonawitz2017practical,elkordy2022much,so2021lightsecagg,secagg_so2021securing,9712310}. Particularly, when a server can only access the updates aggregated across hundreds or thousands of training images, the reconstruction process becomes very challenging. Gradient inversion attacks are impossible without additional model modifications or training rounds. This is where linear layer leakage attacks~\cite{fowl2022robbing,boenisch2021curious} have shown their superiority.

This sub-class of analytic data reconstruction attacks is based on the server crafting maliciously modified models that it sends to the clients. In particular, the server uses a fully-connected (FC) layer to leak the input images. Compared to any other attack, linear layer leakage attacks are the only methods able to scale to an increasing number of clients or batch size, maintaining a high total leakage rate. This is done by continually increasing the size of an FC layer used to leak the images. For example, with 100 clients and a batch size of 64 on CIFAR-100, an attacker can leak $77.2\%$ of all images 
in a single training round using an inserted FC layer of size 25,600. In this case, the number of units in the layer is $4\times$ the number of total images, and maintaining this ratio when the number of clients or batch size increases allows the attack to still achieve roughly the same leakage rate. Despite the potential of linear layer leakage, however, an analysis of the limits of its scalability in FL has been missing till date. 

In this work, we dive into this question and explore the potential of scaling linear layer leakage attacks to secure aggregation. We particularly highlight the challenges in resource overhead corresponding to memory, communication, and computation, which are the primary restrictions of cross-device FL. We discover that while SOTA attacks can maintain a high leakage rate regardless of aggregation size, the overhead is massive. With 1,000 clients and a batch size of 64, maintaining the same leakage rate as before would result in the added layers increasing the model size by 6GB. There would also be an added computation time of 21.85s for computing the update for a single batch (size 64), 
a $10\times$ overhead compared to a baseline ResNet-50. This is a massive problem for resource-constrained FL where clients have limited communication or computation budgets.

However, this problem arises from an incorrect perspective from prior work where they treat the attack on an aggregate update the same as an individual client update. Specifically, we argue that it is critical to treat an aggregation attack not as an attack on a single large update, but as individual client updates combined together. In the context of linear layer leakage, this is the difference between separating the scaling of the attack between batch size and the number of clients or scaling to all images together. 

Following this, we use the attack \name~\cite{our2022mandrake} with sparsity in the added parameters between the convolutional output and the FC layer to highlight the difference in model size compared to prior SOTA. The addition can decrease the added model size by over 327$\times$ and decrease computation time by 3.34$\times$ compared to SOTA attacks while achieving the same total leakage rate. For a batch size of 64 and 1000 clients participating in training, the sparse \name module adds only a little over 18MB to the model while leaking $77.8\%$ of the total data 
in a single training round (comparable to other SOTA attacks). 

We discuss other fundamental challenges for linear layer leakage including the resource overhead of leaking larger input data sizes. We also discuss that sparsity in the client update fundamentally cannot be maintained through secure aggregation and the client still accrues a communication overhead when sending the update back to the server. All other aspects of resource overhead such as communication cost when the server sends the model to the client, computation time, and memory size, are decreased through sparsity.

Our contributions are as follows:
\vspace{-5 pt}
\begin{itemize}[noitemsep]
    \item We show the importance of sparsity in maintaining a small model size overhead when scaling to a large number of clients and the incorrect perspective prior work has had when treating the aggregate update as a single large update. By using sparsity with \name and attacking 1000 clients with a batch size of 64, the added model size is only 18.33 MB. Compared to SOTA attacks, this is a decrease in over $327\times$ in size and also results in a decreased computation time by $3.3\times$ while maintaining the same leakage rate.
    
    \item We show the fundamental challenge of linear layer leakage attacks for scaling attacks towards leaking larger input image sizes and the resulting resource overhead added.
    
    \item We show the problem of maintaining sparsity in secure aggregation when the encryption mask is applied, which adds to the communication overhead when clients send updates back to the server.
\end{itemize}

\section{Related work}
\label{sec:related_work}
We are interested in data reconstruction attacks in the setting of FL under secure aggregation (SA)~\cite{bonawitz2017practical}. Under SA, a server cannot gain access to any individual client's updates. Participating clients encrypt their updates such that only after a server aggregates them will it have access to an unencrypted aggregate update. This section discusses prior work in data reconstruction attacks and their applicability toward this challenging scenario.

\noindent
\textbf{Gradient inversion.} An attacker with access to the model parameters and an honest individual gradient performs a gradient inversion attack by initializing random dummy data and minimizing the difference between the gradient computed by the dummy data and the ground truth gradient. Many works have looked to improve reconstruction through zero-shot label restorations~\cite{yin2021see,zhao2020idlg,geng2021towards}, regularizers~\cite{geiping2020inverting}, or the use of multiple initialization seeds~\cite{yin2021see}. However, they cannot scale to aggregation because the computational complexity scales with an increasing number of images as $\mathcal{O}(n\times dim_{input})$~\cite{kariyappa2022cocktail}, where $n$ is the total number of images. For FL, $n$ is the batch size $\times$ number of clients.

\noindent
\textbf{Analytic attacks.} Analytic attacks involve parameter manipulation~\cite{wen2022fishing,pasquini2021eluding} or attempting to dis-aggregate the gradient by observing multiple training rounds~\cite{lam2021gradient}. While these methods work in the aggregate setting, they are not scalable towards an increasing number of clients. \cite{wen2022fishing} can only attack a single training image within a single round and~\cite{pasquini2021eluding} can only attack a single client. ~\cite{lam2021gradient} can support an increasing number of clients, but requires additional side-channel information not required for FL and additionally can require hundreds or thousands of training rounds to succeed. It also relies on optimization, so if the client batch size is larger, reconstruction quality will diminish. 

\noindent
\textbf{Linear layer leakage.} A sub-class of analytic methods is linear layer leakage attacks~\cite{fowl2022robbing,boenisch2021curious}. These attacks function with an inserted module that is typically two FC layers (linear layers) at the start of the model architecture. The attacks are then able to use the gradients of the first FC layer to directly recover the inputs to the layer. Since the FC layer is placed at the start of the architecture, the inputs are the training images themselves. Specifically, if an image activates a neuron in an FC layer, the image can be reconstructed as 
\vspace{0 pt}
\begin{equation}\label{eq:1}
    x = \frac{\delta L}{\delta W^i} / \frac{\delta L}{\delta B^i}
\vspace{-1 pt}
\end{equation}
\noindent
where $x$ is the recovered image and $\frac{\delta L}{\delta W^i}, \frac{\delta L}{\delta B^i}$ are the weight and bias gradient of the activated neuron~\cite{phong2017privacy}.

These recovered images are near-exact reconstructions. However, if multiple images activate the same neuron, the reconstructed image becomes a combination of these images. Prior work has proposed binning~\cite{fowl2022robbing} and trap weights~\cite{boenisch2021curious} to prevent collision of activated neurons between different images. Trap weights aim to create a sparse activation by initializing the FC layer weights as half negative and positive, with a slightly larger negative magnitude. Under binning, the weights of the FC layer are set such that they measure an aspect of the image, such as the image brightness or pixel intensity. A ReLU activation is used and the neuron biases increase (negatively) so that subsequent neurons allow fewer images to activate them. For any neuron, if only one image has it as the activated neuron with the largest cut-off bias, we can reconstruct the image as
\vspace{-1 pt}
\begin{equation}\label{eq:5}
    x^i = (\frac{\delta L}{\delta W^i} - \frac{\delta L}{\delta W^{i+1}}) / (\frac{\delta L}{\delta B^i} - \frac{\delta L}{\delta B^{i+1}})
\vspace{-1 pt}
\end{equation}
where $x^i$ is the reconstructed image, $i$ is the activated neuron, and $i+1$ is the neuron with the next largest cut-off that was not activated. This method can scale to larger number of clients or batch size while maintaining a high leakage rate by increasing the number of units in the FC layer~\cite{fowl2022robbing,our2022mandrake}. However, this scalability comes at the cost of an increasing model size and becomes much worse under aggregation, as the number of images increases multiplicatively with the batch size and number of clients.

Another similar method uses blind-source separation of an FC layer~\cite{kariyappa2022cocktail} to reconstruct images. This method can support only reconstructions up to 1024 images and, in the context of FL, is a small scale attack and is not particularly applicable to scaling in FL. The size overhead added by the method is not insignificant, as an FC layer added to the start of the model for attacking a batch size of 1024 will be a minimum of a 768MB model size overhead.

The size overhead added by scaling these methods is a fundamental problem. With~\cite{fowl2022robbing,boenisch2021curious}, these methods treat aggregation attacks the same as individual client attacks, evident through the statement that \textit{"given an aggregated gradient update, we always reconstruct as discussed in [the methodology section]"}~\cite{fowl2022robbing}. ~\cite{kariyappa2022cocktail} falls under similar thinking, applying their attack on aggregate updates as simply the same attack on a larger batch size. Another work~\cite{qian2021minimal} discusses how a full batch can be recovered as long as the number of units is larger than the total number of images. While many attacks have not been applied to aggregation yet, it is clear that there is no key difference in the perspective of applying attacks to aggregation compared to individual updates.

Our work is mainly focused on linear layer leakage attacks, but the applicability will be relevant to other methods as they explore large-scale attacks on aggregation. For example, while optimization attacks still do not have a good method of scaling to aggregation due to an increasing computational complexity, a dual problem has been shown where multiple solutions exist for a single update~\cite{zhang2022survey}. If future work discusses model modifications in an increased width or depth of the model to reconstruct larger numbers of images, our work will be relevant. This is also likely since prior work has already begun discussing the relationship between model size/depth and reconstruction ability~\cite{geiping2020inverting}. If computational complexity is not a problem, these same ideas will be used for attack scalability.

In the next section, we will discuss why the prior work perspective on attacking aggregate gradients as a single large batch is a problem and how it leads to large resource overheads in linear layer leakage. We will show how the design of \name with separate scaling between the number of clients and batch size uses the correct perspective and allows us to use sparsity to decrease overhead.
\vspace{-5 pt}
\section{Decreasing the resource cost}
\label{sec:methodology}

\subsection{Requirement for linear layer leakage}
Linear layer leakage relies on a fundamental requirement. Since the images are leaked from the gradients, the images must be able to be stored in the model gradients. For example, consider a CIFAR-100 image ($32\times32\times3$). In order to store the image in the update, the total number of gradients must be at least $32\cdot32\cdot3=3,072$. For a batch of 64 images, this number would then become $3,072\cdot64=196,608$. If aggregation across 100 clients is added, this would then be $3,072\cdot64\cdot100=19,660,800$ total gradients. These gradients come from the weights connecting the input image to the first FC layer, and the minimum size of the first FC layer would need to be $6,400$ units just to be able to store all image information.

However, this only considers the case where images are stored in the gradients perfectly. In reality, multiple images can activate the same neuron causing overlap as discussed in Section~\ref{sec:related_work}, and in order to maintain a high leakage, the number of neurons must be greater than the total number of images. We find experimentally using the binning approach from Robbing the Fed~\cite{fowl2022robbing}, that if the number of neurons is $4$ times the number of images, we can achieve an overall leakage rate of $70-80\%$ on Tiny ImageNet~\cite{le2015tiny}, MNIST~\cite{lecun1998mnist}, and CIFAR-100~\cite{krizhevsky2009learning}. However, with the previous example, this would be over $78.6$ million gradients, where these gradients would come from the weight parameters of an inserted FC layer at the start of the network. Furthermore, another FC layer would be needed to resize the previous FC layer to the input image size prior to input to the rest of the model. This process then adds another $78.6$ million weight parameters, making the total size about $157.3$ million weights, roughly $13.46\times$ the size of a ResNet-18.

From the previous example, we can see the difficulty in scaling linear layer leakage attacks to the FL setting in terms of model size. An increased model size will exacerbate the fundamental problem of FL: the clients have resource restrictions. This additional overhead will affect all aspects of resource constraints, increasing the memory required for storage and training the model along with the communication and computational costs associated. With a larger model size, receiving the model and sending the update back to the server will be more costly for the client. Similarly, a larger model will result in a longer time to compute the update. Our goal then is to minimize these costs.

\subsection{Single client overhead and sparsity}
A batch gradient is the average across all gradients of the individual training samples in the batch. Aggregation is done on top of the batch gradients across multiple clients. Following this, the aggregation of client updates can be interpreted as a single large batch aggregation. This leads to the natural perspective of prior work where an attack on an aggregate update is simply an attack on a single large batch. However, there is a key difference between a large batch update and multiple smaller batch updates being aggregated together to make up a single update when approaching from an attack perspective. Specifically, the storage requirement of linear layer leakage for each individual client is not the same as for all clients combined.

For an individual client with a batch size of 64 on CIFAR-100, an FC layer needs to have $786,432$ weight gradients to maintain a 4:1 ratio of neurons to images. This does not change regardless of how many other clients are present. While this is clear for individual client attacks, the application towards SA is much less obvious. Since an attacking server only has access to the aggregate update, the total number of weight gradients must still be large enough to store all images across all clients. However, our concern is with the resource overhead of individual clients. The prior work perspective of treating the aggregate update as a single large batch means each client must take on the full overhead of the total number of images. Despite this, individual clients only actually need enough for their own. 

Hence, we propose the use of sparsity as the primary method for decreasing the resource overhead of linear layer leakage. Given SA, a server can access only the aggregate update, and as a result, the model must still be large enough to contain all images across participating clients. However, each individual client update only needs to be large enough to support their own images. If all added parameters and gradients are zero outside of this small set used for each individual client's images, the properties still hold. Thus, the added parameters for the entire model are large enough to store all images across clients, but individual clients will only have a small set of non-zero parameters. The size of this small set is also irrespective of the number of clients in the aggregation, only needing to scale to the batch size for each individual client. With a high level of sparsity in the model parameters and updates, sparse tensors can be utilized to decrease the resource overhead. Sparse tensors are representations aimed at the efficient storage of data that is mostly comprised of zeroes. We use the COO (coordinate) format, a common sparse representation in PyTorch~\cite{paszke2019pytorch} that stores the indices and values of all the non-zero elements. When non-zero elements make up a small part of the total size, this leads to more efficient memory usage and quicker computation, both desirable traits for FL. Additional compression can also be used on top of sparsity to further decrease communication costs.

\begin{figure}[!t]
\begin{center}
\includegraphics[width=0.9\columnwidth]{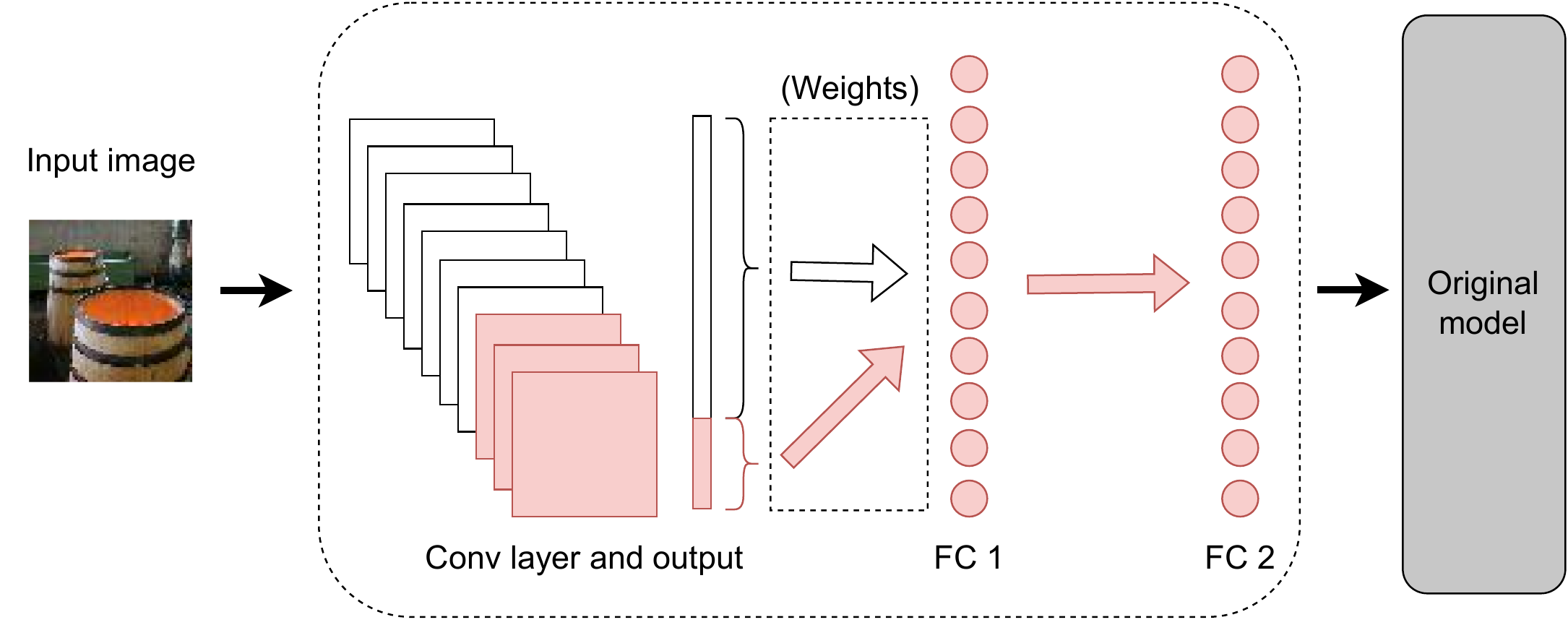}
\end{center}
\vspace*{-5mm}
\caption{\label{fig:mandrake_attack} \name attack allows for sparsity by design. The red color indicates non-zero parameters and the white is zeros. The majority of added parameters come from the weights connecting the convolutional layer output and the first FC layer and only $\frac{1}{N}$ of this is non-zero, where $N$ is the number of clients.}
\vspace*{-5mm}
\end{figure}

\subsection{Convolutional layer for sparsity}
\name encapsulates the idea of producing sparsity within the attack design through the additional placement of a convolutional layer in front of the 2 FC layers used by standard linear layer leakage methods~\cite{our2022mandrake}. Figure~\ref{fig:mandrake_attack} shows an attack overview. An input image to the convolutional layer can be directly passed through using a number of kernels equal to the image channels. Each convolutional kernel will push a different input channel forward using weight parameters of all zeros and a single one in the center of a different kernel channel. For a 3-channel image, only 3 convolutional kernels will be required to push the image through.

The addition of a convolutional layer allows another level of attack scalability in the number of convolutional kernels in the model. Particularly, the number of convolutional kernels is chosen based on the number of color channels in the input images, and this scales with the number of clients attacked. If we have 3-channel input images and 100 clients, $3\cdot100=300$ kernels are used. Each client would use a different set of 3 kernels in the convolutional layer to push their images forward. All other kernel parameters can be set to zero. Similarly, only the weight parameters connecting the output of those 3 non-zero kernels to the FC layer will be non-zero. For this connection between convolutional output and the FC layer, the number of non-zero weight parameters would be
\vspace{-3 pt}
\begin{equation}\label{eq:2}
    |\{w_N \text{ s.t. } w_N\neq0\}| = \frac{1}{N} \cdot |w_N|
\vspace{-3 pt}
\end{equation}
\noindent
where $N$ is the number of clients in aggregation and $|w_N|$ is the total number of weight parameters connecting the convolutional output and FC layer. The number of non-zero weights is also constant regardless of the number of clients
\begin{equation}\label{eq:3}
    |\{w_N \text{ s.t. } w_N\neq0\}| = |\{w_{N+1} \text{ s.t. } w_{N+1}\neq0\}|
\end{equation}
\noindent
For a client batch size of 64 on CIFAR-100, using an FC layer of 256 units results in a leakage rate of $77\%$ using the same binning strategy of~\cite{fowl2022robbing}. The number of non-zero weight parameters between the convolutional output and FC layer would be $(32\cdot32\cdot3)\cdot256=786,432$, only $1\%$ of the number of non-zero parameters compared to prior work. 

Additionally, scaling the convolutional layer means that the FC layer will stay at a fixed size and only scale to the client batch size. This is particularly useful for preventing a size increase in the weights of the second FC layer. Since the second FC layer resizes the output of the previous FC layer (used for leakage) to the size of the input image, naively increasing the size of the first FC layer results in an increase in the same increase in the size of the second. The final design then has these layer sizes. The convolutional layer has $N\times input_{ch}$ kernels, the first FC layer has a number of units equal to batch size $\times$ 4 (the ratio of neurons to images), and the second FC layer has a number of units equal to the input image size. What we then see is that \name has roughly half the total parameters of~\cite{fowl2022robbing} while maintaining the same leakage rate. The number of weight parameters (non-zero and zero) is
\begin{equation}\label{eq:4}
    |w_N| = dim_{input} \cdot N \cdot |\text{FC layer}| + |\text{FC layer}| \cdot dim_{input}
\end{equation}
\noindent
where $dim_{input}$ is the input image size. The $|\text{FC layer}|$ depends on the batch size and does not change regardless of the number of clients. For a batch size of 64, we fix it to be 256 units and achieve a $77\%$ total leakage rate. Increasing or decreasing this layer size further can result in a higher or lower leakage rate and model size respectively.

On the other hand, Robbing the Fed, which achieves prior SOTA in leakage rate and number of additional parameters, adds a total number of $|w_{N,RtF}| = 2\cdot dim_{input} \cdot N \cdot |\text{FC layer}|$ parameters from their method. The value of the size of the FC layer is fixed to be the same as our method in order for comparison. In aggregation, when $N>>1$ we have that $|w_{N,ours}| \approx \frac{1}{2}\cdot |w_{N,RtF}|$. This is considering all zero and non-zero parameters equally for our method. Robbing the Fed does not have non-zero weight parameters, so we also have that $|\{w_{N,ours} \text{ s.t. } w_{N,ours}\neq0\}| \approx \frac{1}{N} \cdot |\{w_{N,RtF} \text{ s.t. } w_{N,RtF}\neq0\}|$.

We did not use the convolutional kernel weights or the biases added during the comparison in the number of parameters, but they are significantly fewer than the weight parameters of the FC layers. For 100 clients, the number of parameters added by the convolutional kernel weights and the layer biases are only $0.01\%$ of the FC layer weights. 

\subsection{Linear layer leakage method}
While the convolutional layer allows for sparsity and separate leakage between each client, the underlying methodology of the linear layer leakage is still important. We use the binning methodology of Robbing the Fed~\cite{fowl2022robbing} instead of the trap weights in~\cite{boenisch2021curious} since the leakage rate achieved with the same FC layer size is higher for binning. 

The approach of using a convolution layer to separate the weight gradients between clients prevents the FC layer from increasing, but also means we cannot retrieve the individual bias gradients $\frac{\delta L}{\delta B^i}$, as they will be aggregated between clients. However, knowing the bias gradient values are not important for reconstruction. If we know the range of values, we can directly scale the weight gradients. If the images are between [0,1], we can recover the images using only the weight gradients through
\begin{equation}\label{eq:6}
    x^i_{k} = \frac{abs({\frac{\delta L}{\delta W^i}_k} - \frac{\delta L}{\delta W^{i+1}}_k)}{max(abs({\frac{\delta L}{\delta W^i}_k} - \frac{\delta L}{\delta W^{i+1}}_k))}
\end{equation}
where we scale the weight gradient such that it has a maximum value of 1. If the ground truth image has a max value of 1, the reconstructed image will be exact. If this is not the case and the maximum is lower, the reconstruction will have a shifted brightness. This approach is described further in~\cite{our2022mandrake} and the images are easily identifiable after the range shift. This process does not cause issues with reconstructions either, with the method still achieving a high SSIM~\cite{wang2004image} and L-PIPS score~\cite{zhang2018unreasonable}.

\subsection{Secure aggregation masking}
While sparsity allows us to take advantage of the large number of zero parameters in the model, the property becomes difficult to maintain through SA, as a non-sparse mask will be used regardless of whether the individual client update is sparse or not. 
Thus, even though the client updates are sparse, SA applies a non-sparse mask on top of the update such that it is encrypted. Since masking removes the property of sparsity from the update, the client incurs a communication overhead when sending the update back to the server which will not be mitigated. For 100 clients with a batch size of 64 on CIFAR-100, the model size that is transmitted from the server to the client by \name is 18.04MB and the update sent back to the server is 303.33MB. Robbing the Fed is larger than both, adding a size overhead of 600.11MB to both ends of communication.

However, while sparsity does not benefit the communication cost when the client sends the update back to the server, it benefits all other aspects of client resource overhead, including when the server sends the model to the client, the storage on the client, and the time for computing the update. The total number of added parameters of \name is also half the size of Robbing the Fed.

\subsection{Broad applicability of sparsity}

Sparsity can help with many forms of attacks with FL. While we use the binning method of~\cite{fowl2022robbing}, sparsity also helps the trap weight methodology~\cite{boenisch2021curious} differently. We find that the baseline attack of trap weights is unable to scale to an increasing number of clients in aggregation. As the total number of images increases, even if the ratio of neurons to images remains the same, the leakage rate will decrease (we refer to the supplement for experiments). However, using the convolutional layer method of \name, the leakage process is separate between clients. This will prevent the leakage rate decrease with an increasing number of clients.

While we previously explored the application of sparsity in linear layer leakage attacks, the idea can be applied to other attacks when scaling to aggregation. For example, sparsity can be used in the same way for the blind-source separation method of Cocktail Party attack~\cite{kariyappa2022cocktail} when scaling to aggregation. This would result in both model size and computation complexity decrease. Using the original method of Cocktail Party, the complexity would be $\mathcal{O}(n\times n)$~\cite{kariyappa2022cocktail}, where $n$ is the total number of images. However, using sparsity would decrease the computational complexity by lowering $n$ from the total number of images to just the batch size of the individual client instead.

Along the same line, sparsity could be brought to gradient inversion to decrease the computational complexity. The original challenge in scaling to aggregation for gradient inversion is that the number of total images is significantly larger. However, sparsity once again can be used to decrease the computational complexity $\mathcal{O}(n\times dim_{input})$, so that $n$ is the client batch size instead of the total number of images. This approach would require model modification to introduce sparsity similar to \name, resulting in a model size increase. However, the storage size benefits of sparsity can also help decrease the overhead.

\vspace{-5 pt}
\section{Experiments}
\label{sec:expeirments}
\begin{figure}[!t]
\vspace*{-0.5mm}
\begin{center}
\includegraphics[width=1.0\columnwidth,trim={0 0 0 10mm},clip]{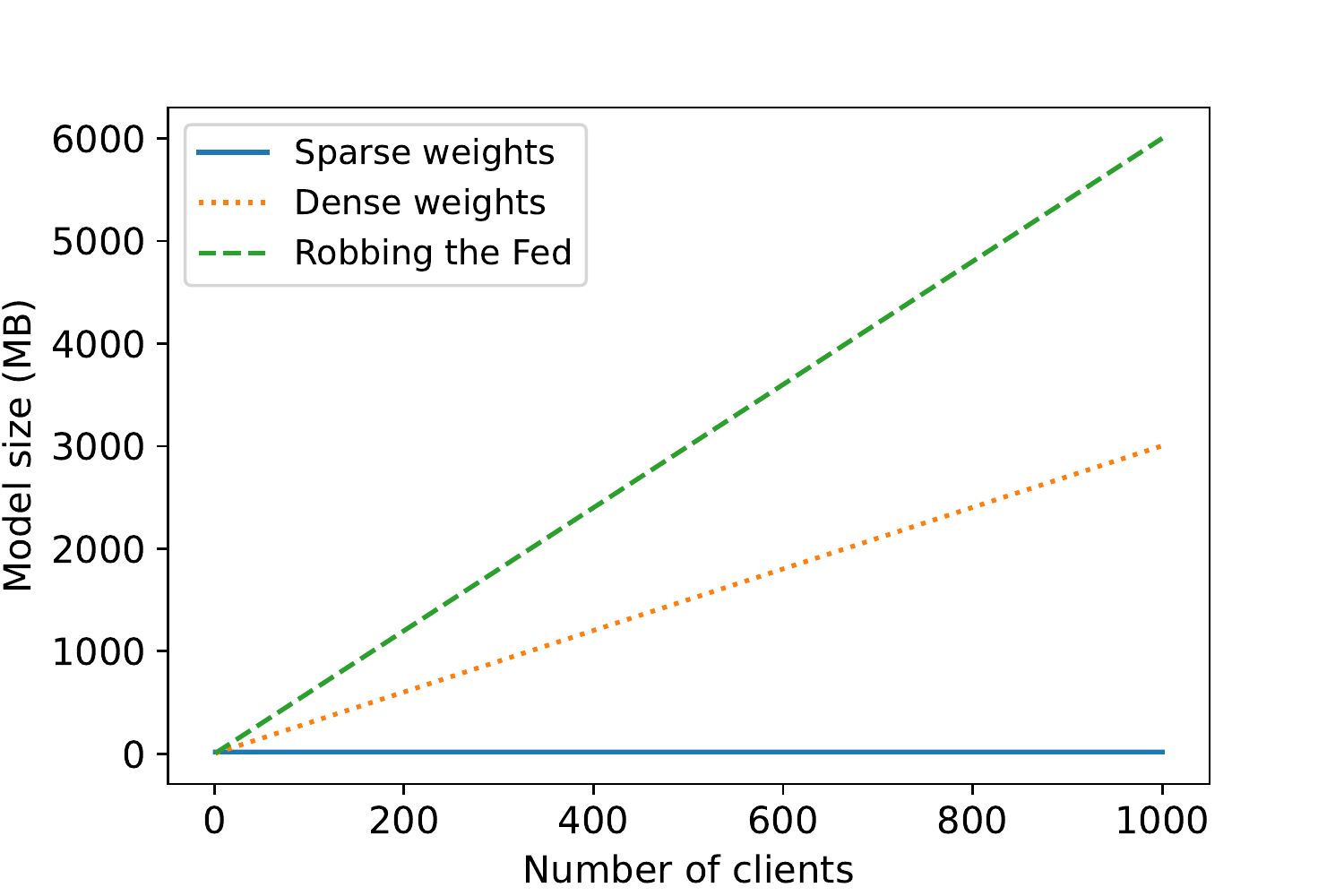}
\end{center}
\vspace*{-5mm}
\caption{\label{fig:model-size-all} Comparing the model size of the dense and sparse tensor attack with Robbing the Fed on the downsampled Tiny ImageNet dataset with a client batch size of 64. The model sizes are given when achieving a total leakage rate of $77\%$. At 1000 clients, the sparse representation is $327.33\times$ smaller than Robbing the Fed.}
\vspace*{-5 mm}
\end{figure}


We evaluate our attack in the secure aggregation FL setting. We are particularly focused on the resource costs in terms of model size and computation overhead added by linear layer leakage attacks when scaling to larger numbers of clients. We primarily compare three attacks: our attack using dense tensors, our attack using the sparse tensor representation, and Robbing the Fed~\cite{fowl2022robbing} the prior SOTA linear layer attack. For all experiments, we use a client batch size of 64 with a varied number of clients in aggregation. We use a $|\text{FC layer}|$ $4\times$ the number of images. Using the binning method of ~\cite{fowl2022robbing}, both our method and Robbing the Fed achieve the same total leakage rate on the Tiny ImageNet dataset~\cite{le2015tiny}. Using a single training round with 1000 clients, \name leaks $76.9\%$ (49,209) images, and Robbing the Fed leaks $76.5\%$ (48,992) of the total 64,000 images. For additional examples of reconstructed images and the leakage rate for other datasets, we refer to the supplementary material.

We first show the model size comparison for each of the methods on a downsampled Tiny ImageNet dataset ($32\times32\times3$), comparing the model size trend with increasing clients. We show the overhead added to standard vision models from each method. In Section~\ref{sec:experiments-larger-ims} we also look at the size overhead of the inserted modules on the MNIST ($28\times28\times1$)~\cite{lecun1998mnist}, CIFAR-100 ($32\times32\times3$)~\cite{krizhevsky2009learning}, Tiny ImageNet ($64\times64\times3$), and ImageNet ($256\times256\times3$)~\cite{russakovsky2015imagenet} datasets. We use this section to highlight the difficulties in scaling with larger input image sizes.

When using larger input image sizes with a large number of clients, the FC layer size of Robbing the Fed grows too large for memory. As a result, we use the downsampled Tiny ImageNet dataset for these comparisons. We run the attacks on a CPU compared to a GPU, focusing on the resource restrictions of cross-device FL. For the computation overhead, we compare the additional time required to compute the model gradients when compared to a baseline ResNet-50~\cite{he2016deep} from PyTorch. We place the extra layers at the start of the architecture.

Finally, we experimentally show the binning methodology of Robbing the Fed~\cite{fowl2022robbing} is more effective than trap-weights~\cite{boenisch2021curious} in terms of mutual information. 

\begin{table}[!t]
\begin{center}
\small
\begin{tabular}{|l|c|cc|}
\hline
                                                                       & \textbf{\begin{tabular}[c]{@{}c@{}}Model size \\ (MB)\end{tabular}} & \textbf{\begin{tabular}[c]{@{}c@{}}Sparse \\ attack\end{tabular}} & \textbf{\begin{tabular}[c]{@{}c@{}}Robbing \\ the Fed\end{tabular}} \\ \hline
\textbf{MobileNet v3 (L)} & 20.9161                                                             & 87.65\%                                                            & 28690.76\%                                                           \\
\textbf{ResNet-18}                                                     & 44.5919                                                             & 41.11\%                                                            & 13457.57\%                                                            \\
\textbf{ResNet-50}                                                     & 97.4923                                                             & 18.80\%                                                            & 6155.35\%                                                             \\
\textbf{Inception v3}                                                  & 103.6120                                                            & 17.69\%                                                            & 5791.79\%                                                             \\
\textbf{VGG-11}                                                        & 506.8334                                                            & 3.62\%                                                            & 1184.02\%                                                             \\ \hline
\end{tabular}
\end{center}
\vspace*{-4mm}
\caption{\label{tab:model-size-comparison} Model size overhead added from the attacks with 1000 clients and a batch size of 64 on Tiny ImageNet compared to vision models. The overhead added by the sparse representation attack ($18.33$MB) is significantly smaller than Robbing the Fed ($6000.99$MB) and achieves the same leakage rate.}
\vspace*{-3mm}
\end{table}
\vspace{-4 pt}
\subsection{Model size}
We start with a discussion on the model size. For these experiments, we use PyTorch's sparse COO (Coordinate format) tensor representation~\cite{paszke2019pytorch}. This format stores the non-zero values in indices and values tensor. The size of the sparse tensor in bytes is
\vspace{-4 pt}
\begin{equation}\label{eq:7}
    size = (dim \cdot 8 + data size) \cdot |\{w_N \mid w_N\neq0\}|
\vspace{-4 pt}
\end{equation}
following PyTorch's sparse tensor memory consumption. The tensor dimensions are $dim=2$ and the data size is $4$ bytes for our model.

When the ratio of the number of neurons to images is $4:1$, the attack methods achieve $~77\%$ total leakage rate on the Tiny ImageNet dataset (small randomness coming from batch images selection). In the case of Robbing the Fed, this is achieved when the $|\text{FC layer}| = (\text{num clients})\cdot(\text{batch size})\cdot 4$. Our method achieves this with a fixed $|\text{FC layer}|=256$ by increasing the number of convolutional kernels by 3 for each client.

Figure~\ref{fig:model-size-all} shows the model size overhead (MB) added by the 3 methods with a fixed leakage rate and a varying number of clients. At 3 clients, the sparse representation is nearly the same size as Robbing the Fed ($99.994\%$). At 5 clients, the sparse and dense (update size with SA sent back to server) representations are the same size. As the number of clients grows, both dense weights and Robbing the Fed quickly grow in size, while the sparse representation remains virtually the same. While the method of Robbing the Fed is able to achieve the same total leakage rate, the number of parameters is roughly double the dense weights attack. With 1000 clients, Robbing the Fed is $327.33\times$ larger than the sparse tensor attack. Between $1-1000$ clients, the size overhead of the sparse representation increases from 18.04MB to 18.33MB. The small size increase comes from the convolutional kernel parameters and biases.

Table~\ref{tab:model-size-comparison} shows the percentage overhead added by the sparse tensor attack and Robbing the Fed on several standard vision models. There are 1000 clients in aggregation with a batch size of 64. The sparse tensor representation adds a significantly smaller overhead ($18.33$MB) compared to Robbing the Fed ($6000.99$MB) while achieving the same leakage rate. Even with a large model like VGG-11, Robbing the Fed adds a massive model size overhead increase of $1184.0\%$, while the sparse attack only adds $3.6\%$.

We note that using a compressed sparse row (CSR) tensor representation results in a model size overhead of only roughly $\frac{2}{3}$ compared to the COO representation. At 1000 clients, the size added using sparse CSR is only $12.33$MB. However, this sparse tensor representation is currently in the beta phase of PyTorch, so we only use sparse COO tensors for the experimental comparisons. 

\begin{figure}[!t]
\vspace*{-2.5mm}
\begin{center}
\includegraphics[width=1.0\columnwidth,trim={2mm 0 2mm 10mm},clip]{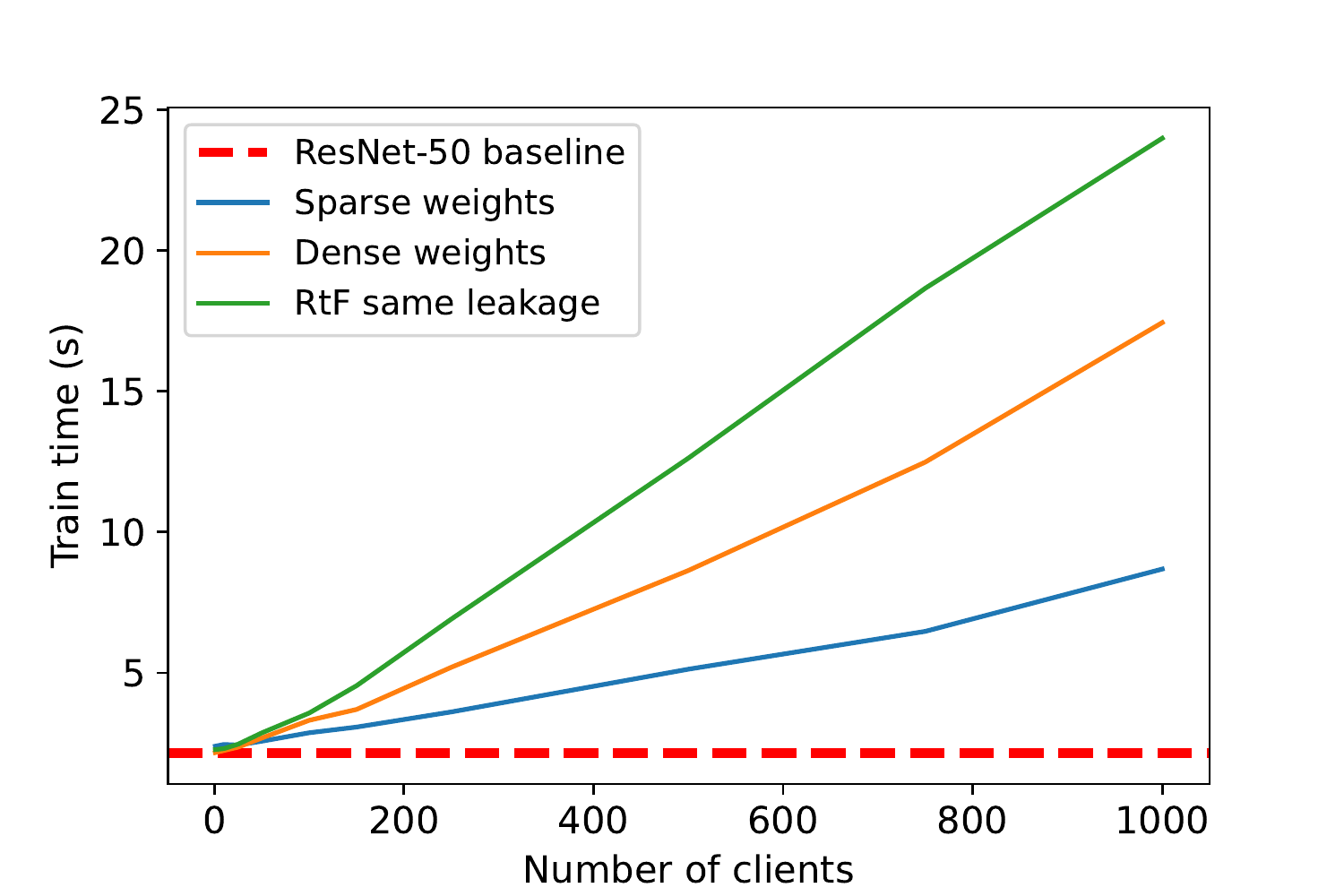}
\end{center}
\vspace*{-5mm}
\caption{\label{fig:resnet50-comp} Computational overhead in training time added to a ResNet-50 on a CPU from attacks. At 1000 clients, the sparse tensor method adds a 6.5s overhead while Robbing the Fed adds a 21.8s overhead.}
\vspace*{-5mm}
\end{figure}

\subsection{Computation overhead}
We compare the computational overhead added by the linear layer leakage attacks through a comparison of the time to compute an update for an individual client. This includes the time for a forward pass, loss computation, and gradient computation on a client batch. The baseline model we use is a ResNet-50. The vanilla model uses 2.14 seconds for the update computation on a batch of 64 images.

Figure~\ref{fig:resnet50-comp} shows the time required for the update computation for all three attacks with a varying number of clients and a batch size of 64. With 100 clients, using sparse weights adds a $34\%$ (0.73s) time overhead, dense weights adds a $55\%$ (1.17s) overhead, and Robbing the Fed adds a $67\%$ (1.43s) overhead. At 1000 clients, the overhead is $305\%$ (6.54s), $714\%$ (15.30s), and $1019\%$ (21.85s) respectively. With 1000 clients, the sparse attack adds $3.34\times$ less computational overhead compared to Robbing the Fed.

Much work is going into sparse matrix/tensor optimization~\cite{bell2008efficient, williams2007optimization, dalton2015optimizing, zhao2018bridging}. While these experiments give a brief snapshot of the potential computational differences between methods, we note that as sparse tensor implementations improve, the computation overhead of the sparse weights will continue to decrease.

\begin{table}[!t]
\footnotesize
\begin{center}
\begin{tabular}{|l|c|ccc|}
\hline
                                                                                            & \textbf{Clients} & \textbf{\begin{tabular}[c]{@{}c@{}}Robbing\\ the Fed\end{tabular}} & \textbf{\begin{tabular}[c]{@{}c@{}}Dense\\ weights\end{tabular}} & \textbf{\begin{tabular}[c]{@{}c@{}}Sparse\\ weights\end{tabular}} \\ \hline
\multirow{2}{*}{\textbf{\begin{tabular}[c]{@{}l@{}}MNIST\\ (28x28x1)\end{tabular}}}         & 100              & 153.2                                                             & 77.3                                                            & 4.6                                                              \\
                                                                                            & 1000             & 1532.2                                                            & 766.4                                                           & 4.6                                                              \\ \hline
\multirow{2}{*}{\textbf{\begin{tabular}[c]{@{}l@{}}CIFAR-100\\ (32x32x3)\end{tabular}}}     & 100              & 600.1                                                             & 303.0                                                           & 18.0                                                             \\
                                                                                            & 1000             & 6001.0                                                            & 3003.3                                                          & 18.3                                                             \\ \hline
\multirow{2}{*}{\textbf{\begin{tabular}[c]{@{}l@{}}Tiny ImageNet\\ (64x64x3)\end{tabular}}} & 100              & 2400.1                                                            & 1212.1                                                          & 72.1                                                             \\
                                                                                            & 1000             & 24001.0                                                           & 12012.4                                                         & 72.4                                                             \\ \hline
\multirow{2}{*}{\textbf{\begin{tabular}[c]{@{}l@{}}ImageNet\\ (256x256x3)\end{tabular}}}    & 100              & 38400.9                                                           & 19392.8                                                         & 1152.8                                                           \\
                                                                                            & 1000             & 384001.7                                                          & 192193.1                                                        & 1153.1                                                           \\ \hline
\end{tabular}
\end{center}
\vspace*{-5mm}
\caption{\label{tab:input-image-size} Comparison of model size overhead (MB) using different datasets with batch size 64 and 100 and 1000 clients. At 1000 clients on ImageNet, the sparse representation adds a 1.1GB overhead while Robbing the Fed adds 375GB.}
\vspace*{-6mm}
\end{table}

\subsection{Larger image sizes}
\label{sec:experiments-larger-ims}
\vspace{-1mm}
We revisit the model size to show the overhead added for different image sizes. As discussed in Section~\ref{sec:methodology}, the fundamental requirement of linear layer leakage is to be able to store all image pixels in the gradients. As a result, the input image size directly ties to the model overhead added by the attack. Table~\ref{tab:input-image-size} shows the overhead added from the dense and sparse tensor representation attacks along with Robbing the Fed on several different input image sizes. Results are shown for 100 and 1000 clients with a batch size of 64.

As the input image size increases, so does the size overhead from the inserted module. This size increase trend is (near) directly proportional to the change in input image size. For example, the difference in image size between Tiny ImageNet and ImageNet is $(256\cdot256\cdot3) / (64\cdot64\cdot3)=16$. We see that the size overhead difference for Robbing the Fed is also $38,400.9/2400.1\approx 16$. This scaling property also exists with dense and sparse tensor representations.

The model size overhead added, particularly for the larger image sizes, is extremely large. For Robbing the Fed and the dense weight representation, for 1000 clients on ImageNet, the size overhead reaches 375GB and 188GB respectively. By comparison, the sparse tensor setting is much better for attack scalability, creating a little over 1GB in size overhead for 1000 clients.

These experiments highlight a problem with the model size overhead for current linear layer leakage methods when working with larger input sizes. The need to store image pixels in the gradients means that larger images inherently create larger size overheads. This in turn results in overheads in all aspects of memory, communication, and computation for the clients, and practically, these overheads are too large for FL. For the malicious server, one solution would be to use pooling operations prior to leaking the images. While this method will result in reconstructing downsampled images, leaking full-resolution large-sized images, especially with aggregation, is unrealistic. This fundamental limitation applies to all current linear layer leakage methods. Sparsity can significantly decrease the size overhead, but once the input images become large enough the attacks become infeasible on reasonable-sized devices.

\vspace*{-1mm}
\subsection{Leakage in terms of mutual information}
\vspace{-1mm}
We focus on the differences between binning~\cite{fowl2022robbing} and trap-weights~\cite{boenisch2021curious} in terms of mutual information using the neural estimator proposed in~\cite{belghazi2018mine}. Compared to leakage rate which only considers the number of reconstructed images, the mutual information ratio is a finer-grained metric --- it also captures the information leakage that cannot be reconstructed directly into individual images due to images activating the same neurons and thus ignored by leakage rate.

\begin{figure}[!t]
\vspace*{-7mm}
\begin{center}
\includegraphics[width=1.0\columnwidth,trim={0 0 0 0},clip]{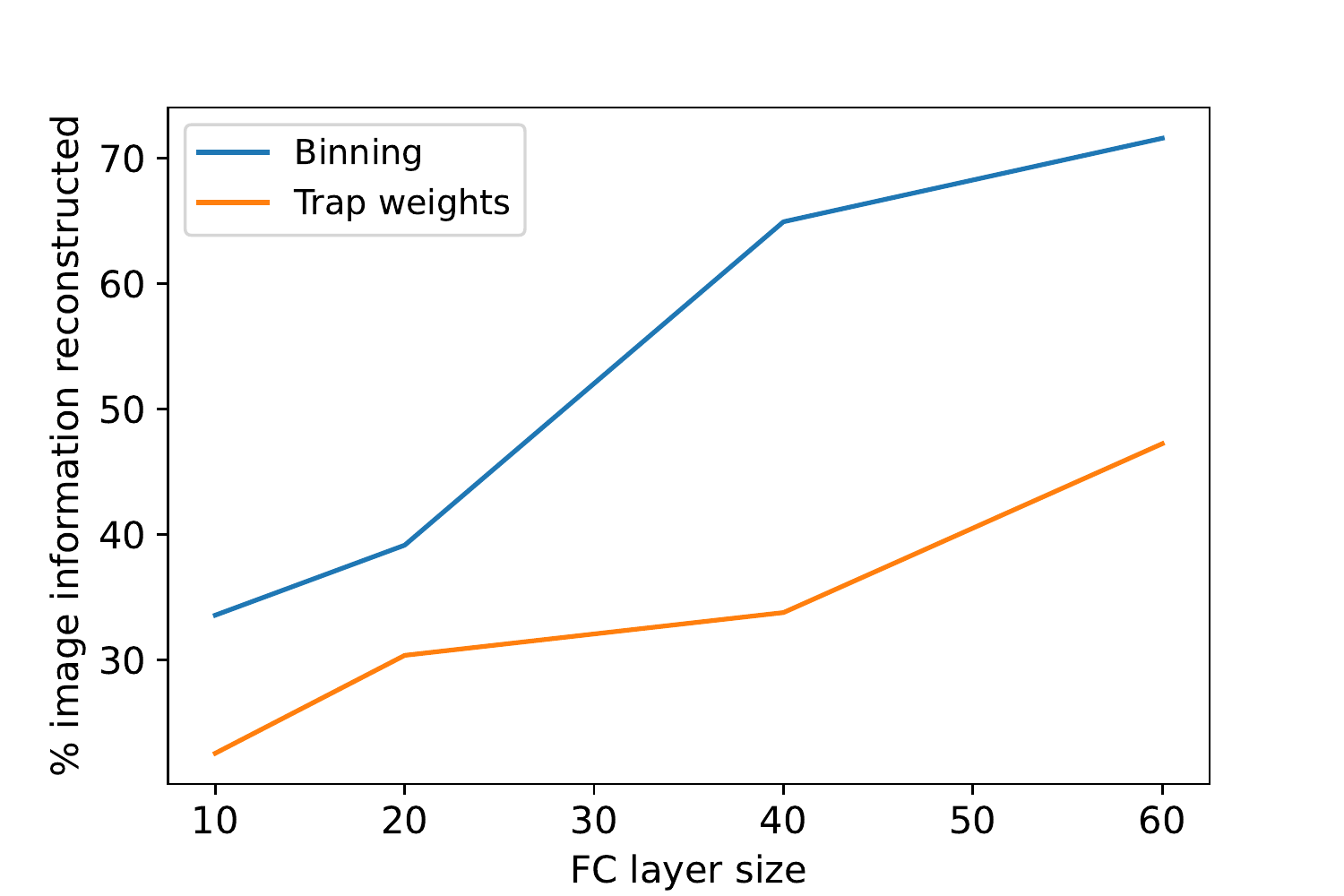}
\end{center}
\vspace*{-6mm}
\caption{\label{fig:MI_leakage_ratio} Comparison of the percentage of the information leaked into the gradient $I(x^{input}_k;g)$ that is recovered through reconstruction $I(x^{input}_k;x_k)$ based on the FC layer size when using binning and trap weights. MNIST dataset with a batch size = 10 is used.}
\vspace*{-6mm}
\end{figure}
Figure~\ref{fig:MI_leakage_ratio} shows that the power of the image reconstruction increases for both the binning and trap weights in terms of the percentage of leaked information as the FC layer size increases. Figure~\ref{fig:MI_leakage_ratio} also shows that for all FC layer sizes, the leakage from trap weights~\cite{boenisch2021curious} is lower than binning~\cite{fowl2022robbing}.

\vspace{-15 pt}

\vspace{-5 pt}
\section{Conclusions}
\label{sec:conclusions}
\vspace{-1mm}
We discuss the fundamental perspective problem of prior work in developing privacy attacks against FL when secure aggregation is used. Attacking the aggregate update as a single large-batch leads to unnecessary resource overheads incurred by clients. By treating the aggregate update as an aggregation of individual client updates we can use parameter sparsity, decreasing the model size by $327\times$ and the computation time by $3.3\times$ compared to SOTA while maintaining the same leakage rate even through SA. We also show the challenge of maintaining sparsity through SA when the client sends the update back to the server and of scaling and leaking large input image sizes. 


\smallbreak
\noindent \textbf{Acknowledgements.} This work was supported by Army Research Lab under Contract No. W911NF-2020-221, National Science Foundation CNS-2038986, Defense Advanced Research Projects Agency (DARPA) under Contract No. HR001120C0156, ARO award W911NF1810400, and ONR Award No. N00014-16-1-2189. Any opinions, findings, and conclusions or recommendations expressed in this material are those of the authors and do not necessarily reflect the views of the sponsors.

{\small
\bibliographystyle{ieee_fullname}
\bibliography{egbib}
}

\appendix
\twocolumn
\section{Leakage rate and reconstructions}
Table~\ref{tab:leakage-rate} gives the leakage rate on the MNIST~\cite{lecun1998mnist}, CIFAR-100~\cite{krizhevsky2009learning}, and Tiny ImageNet~\cite{le2015tiny} datasets using sparse \name and Robbing the Fed~\cite{fowl2022robbing} as discussed in the main paper. We use a batch size of 64 with 100 clients, and the ratio of FC size to batch size is 4:1 (256 unit FC layer). The leakage rate on CIFAR-100 and Tiny ImageNet are roughly the same for both methods. Sparse \name~\cite{our2022mandrake} has a slightly lower leakage rate than Robbing the Fed on MNIST.

\begin{figure*}
     \centering
     \begin{subfigure}[b]{0.49\textwidth}
         \centering
         \includegraphics[width=0.97\textwidth]{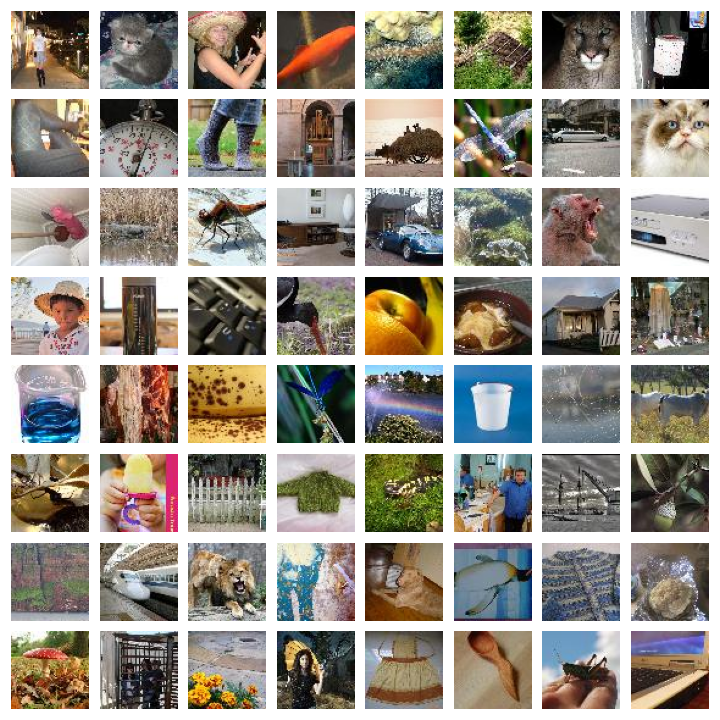}
         \caption{Ground truth}
         \label{fig:ground-truth-tiny-image}
     \end{subfigure}
     \hfill
     \begin{subfigure}[b]{0.49\textwidth}
         \centering
         \includegraphics[width=0.97\textwidth]{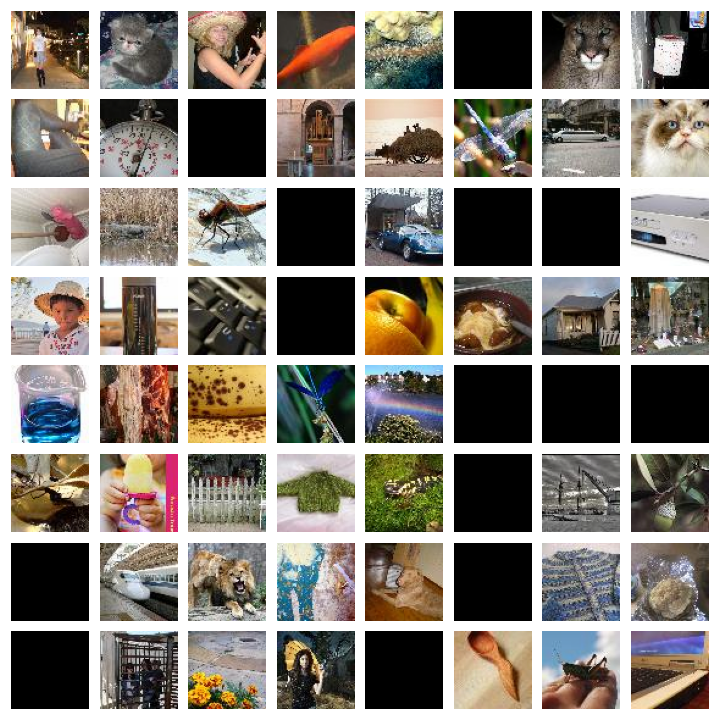}
         \caption{Reconstruction}
         \label{fig:reconstruction-tiny-image}
     \end{subfigure}
\vspace*{-1mm}
\caption{\label{fig:single_client_tinyimagenet} Reconstructed images from Tiny ImageNet from a random client with a batch size of 64 using \name. The ground truth images (a) are shown on the left and the reconstructed images (b) are shown on the right. Any empty boxes within the reconstructed images indicate that reconstruction failed due to an overlap of image activations.}
\end{figure*}

Figure~\ref{fig:single_client_tinyimagenet} shows the ground truth and reconstructions for a single, random client with a batch of 64 on Tiny ImageNet using the sparse \name attack. 50 images were leaked from the client.

\begin{table}[!t]
\begin{center}
\begin{tabular}{|l|cc|}
\hline
                       & \textbf{\begin{tabular}[c]{@{}c@{}}Sparse \\ \name \end{tabular}} & \textbf{\begin{tabular}[c]{@{}c@{}}Robbing\\ the Fed\end{tabular}} \\ \hline
\textbf{CIFAR-100}     & 77.5\% (4957)                                                              & 77.1\% (4931)                                                             \\
\textbf{MNIST}         & 71.0\% (4546)                                                              & 75.1\% (4803)                                                             \\
\textbf{Tiny ImageNet} & 77.8\% (4978)                                                              & 77.7\% (4970)                                                             \\ \hline
\end{tabular}
\end{center}
\vspace*{-4mm}
\caption{\label{tab:leakage-rate} Total leakage rate of sparse \name and Robbing the Fed on various datasets. For all three datasets, 100 aggregated clients and batch size of 64 were used (6400 total images).}
\vspace*{-5mm}
\end{table}

\section{Trap weights under FL}
\begin{figure}[!t]
\begin{center}
\includegraphics[width=1.0\columnwidth,trim={3mm 0 3mm 10mm},clip]{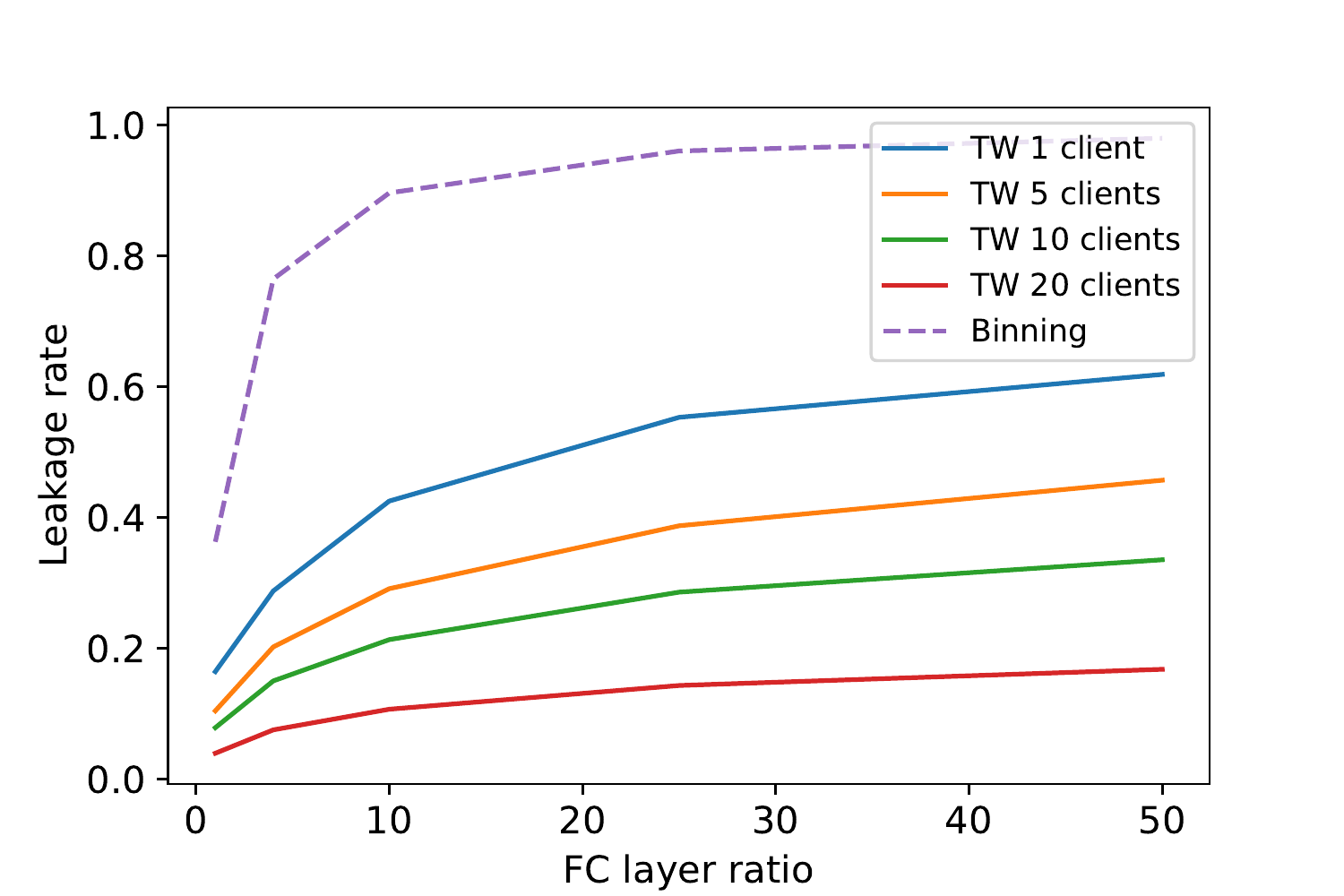}
\end{center}
\vspace*{-6mm}
\caption{\label{fig:wtc-comparison} Leakage rate using trap weights (TW) for a batch size of 64 on Tiny ImageNet and varying the FC layer size ratio and number of clients. The leakage rate decreases with an increasing number of clients even if the ratio of FC size to total number of images remains the same. Binning (Robbing the Fed) has a higher leakage rate at all scales of FC size.}
\vspace*{-4 mm}
\end{figure}

We show the leakage rate using the trap weights attack~\cite{boenisch2021curious} for different FC layer sizes on the downsampled Tiny ImageNet (32x32x3) dataset. We tune the scaling factor between 0.90 and 0.99 (step size of 0.01) to find the highest leakage rate, vary the FC layer ratio (FC layer size = batch size $\times$ num. clients $\times$ FC size ratio), and report the average over 10 runs. We apply this on several numbers of clients and Figure~\ref{fig:wtc-comparison} shows the results compared to binning~\cite{fowl2022robbing}, which has roughly the same leakage rate regardless of the number of clients. Even while maintaining the same ratio between the FC layer size and total number of images, the leakage rate when using trap weights decreases as the number of clients increases.

We note that by using sparsity, trap weights are able to overcome this scalability problem. However, since the binning method of Robbing the Fed achieves a higher leakage rate for all FC layer size ratios, it is still a better choice.

\section{Sparse variant of Robbing the Fed}
The sparse variant of Robbing the Fed (RtF)~\cite{fowl2022robbing} is a method introduced in addition to their baseline in order to apply the attack in the FedAVG setting. The "sparsity" mentioned in Section 4.3 of the RtF paper is discussing how to create activations in the fully-connected (FC) layer such that images should only activate a single neuron instead of a set of neurons. However, this does {\em not} reduce the resource usage added from the attack, which is what we address. With the main change being in the activation function, the same fundamental method as the baseline is used with aggregated updates and the FC layer size still needs to scale to compensate for the total number of images. These layers added to the model will still be fully dense with non-zero parameters.

\section{Evaluating information leakage using mutual information}
In practice, the amount of leaked information is typically quantified as the number of images a malicious server reconstructs (leaked). However, the reconstructions from the attack module can also leak some additional information that is not counted in the leakage rate. For example, while reconstructions of images can overlap, an observer can still obtain information about the training data (e.g., a malicious server who sees an overlap of digits 2, 3, and 8 might be able to identify that an 8 is in the reconstruction). In Section 4, we compared how much information was leaked to the server under a varying FC layer size using either the binning and trap weights method of linear layer leakage attacks, since \name is able to use both.

We used the MNIST dataset for these experiments, and in order to measure the amount of information leaked into the gradient and the amount of information the server was able to reconstruct out of it, we compare the mutual information between: (1) the data batch $x^{input}_k$ at user $i$ and the aggregate gradient $g$ of the attack at the server; (2) the data batch $x^{input}_k$ and the reconstructions $x_k$ at the server for user $k$. Note that by the data processing inequality, we have that:
\begin{equation}\label{eq:ratio}
    \frac{I(x^{input}_k;x_k)}{I(x^{input}_k;g)} \leq 1.    
\end{equation}
since the leaked images were reconstructed only using the gradient. In order to compute the mutual information terms in~\eqref{eq:ratio}, we use the Mutual Information Neural Estimator (MINE) which is the SOTA method~\cite{belghazi2018mine} to estimate the mutual information between two random vectors. For each FC layer size, we sampled 20,000 random batches of the users' data and used each to compute the aggregate gradient $g$ and reconstructed images for a single user $i$. These 20,000 samples were used by MINE to estimate mutual information.

This same procedure was repeated multiple times in order to get multiple mutual information estimates and the average ratio was reported.

\section{FedAVG}
Unlike the gradients of the FC layers, the gradients of the convolutional layer are not necessary for the data reconstruction attack. A malicious server can then send a maliciously crafted model which would freeze the parameters of the convolutional layer to prevent changes from occurring over the local iterations of FedAVG.

\end{document}